\documentclass[letterpaper, 10 pt, conference]{ieeeconf}  

\IEEEoverridecommandlockouts                              

\usepackage{graphics} 
\usepackage{epsfig} 
\usepackage{times} 
\usepackage{amsmath} 
\usepackage{amssymb}  
\usepackage[per-mode=symbol]{siunitx}
\usepackage[dvipsnames,table,xcdraw]{xcolor}
\usepackage{float}

\newcommand{\MDP}{(\mathcal{S}, \mathcal{A}, \mathcal{T}, r, \gamma_t)}

\newcommand{\s}{\ensuremath{\mathbf{s}}}
\newcommand{\g}{\ensuremath{\mathbf{g}}}
\newcommand{\q}{\ensuremath{\mathbf{q}}}
\renewcommand{\v}{\ensuremath{\mathbf{v}}}
\renewcommand{\b}{\ensuremath{\mathbf{b}}}
\newcommand{\R}{\ensuremath{\mathbb{R}}}
\newcommand{\DRS}{\ensuremath{R}}
\newcommand{\PBRS}{\ensuremath{P}}

\usepackage[normalem]{ulem}  

\usepackage{caption}
\usepackage{subcaption}
\usepackage{booktabs}

\usepackage[backend=biber,
    style=ieee,
    bibstyle=ieee,  
    sortcites=true,   
    mincitenames=1,
    maxcitenames=20,
    maxnames=20,
    giveninits=true,            
    uniquelist=false,
    backref=false]{biblatex}
\title{\LARGE \bf
Benchmarking Potential Based Rewards \\ for Learning Humanoid Locomotion
}

\author{Se Hwan Jeon$^{1}$, Steve Heim$^{1}$, Charles Khazoom$^{1}$, and Sangbae Kim$^{1}$
\thanks{$^{1}$All authors are with the Biomimetic Robotics Lab, MIT {\tt\small\{sehwan, sheim, ckhaz, sangbae\}@mit.edu}.}
\thanks{This work was supported by Disney Research Imagineering, NSERC, and the Swiss National Science Foundation (Grant No P2SKP2\_194954).}}

\begin{document}

\maketitle
\begin{abstract}

The main challenge in developing effective reinforcement learning (RL) pipelines is often the design and tuning the reward functions.
Well-designed shaping reward can lead to significantly faster learning.
Naively formulated rewards, however, can conflict with the desired behavior and result in overfitting or even erratic performance if not properly tuned.
In theory, the broad class of \emph{potential based reward shaping} (PBRS) can help guide the learning process without affecting the optimal policy.
Although several studies have explored the use of potential based reward shaping to accelerate learning convergence, most have been limited to grid-worlds and low-dimensional systems, and RL in robotics has predominantly relied on standard forms of reward shaping.
\par
In this paper, we benchmark standard forms of shaping with PBRS for a humanoid robot. 
We find that in this high-dimensional system, PBRS has only marginal benefits in convergence speed.
However, the PBRS reward terms are significantly more robust to scaling than typical reward shaping approaches, and thus easier to tune.

\end{abstract}
\section{INTRODUCTION}

Designing effective reward functions is an iterative process in optimal control pipelines, both when using model-based methods such as model-predictive control (MPC) or model-free methods such as reinforcement learning (RL)~\cite{Xie_LearningLocomotion, marco2016automatic}.
A simple translation of the engineer's intent into a computable function, such as a quadratic error from the desired state or a boolean on task success, often results in unexpected and undesired behavior~\cite{Randlov_BicycleRewardShaping, muller2017quadratic}, especially in RL.
Even when the chosen reward function would yield the desired optimal controller, it can often result in slow convergence and local minima. 
These challenges are typically addressed with \emph{reward shaping}: additional reward terms are added to provide an informative signal of how ``close'' a trajectory is to an optimal policy.
However, defining what ``close'' means in this context is often not intuitive. In practice, a substantial amount of time is spent tuning these shaping rewards to find an acceptable trade-off between convergence and how closely it represents the desired behavior.
Moreover, the entire training process is sensitive to hyperparameters and reward weights, obscuring the effects of a particular reward term on the converged policy and making them challenging to tune precisely~\cite{Andrychowicz_WhatMattersInRL, Xie_LearningLocomotion}.
\par
\textcite{Ng_PolicyInvariance} discuss the special class of \emph{potential-based reward shaping} functions (PBRS) that, in theory, do not affect the final policy.
This theoretical property is highly attractive since PBRS has the potential to decouple the challenges of reward design, allowing the engineer to use simple, task-based rewards to express intended behavior and PBRS to aid convergence.
Several subsequent studies have investigated using PBRS in RL~\cite{Malysheva_RunWithPBRS, Harutyunyan_ArbitraryPBRS, Devlin_DynamicPBRS, westenbroek2022lyapunov} and reported faster convergence, though these studies are typically limited to grid-worlds or low-dimensional systems.
In practice, however, most successful cases of RL in robotics have relied on simpler, direct reward shaping functions that do not enjoy policy invariance properties~\cite{Miki_RLAnymal, rai2018BOBiped, Xie_LearningLocomotion, siekmann2021sim, Li_RobustRLCassie}. 
\par
We present an empirical case study of deep RL for running on a humanoid biped robot in 3D, and systematically compare standard reward shaping and PBRS.
Unlike many previous studies, we find that the main advantage of the PBRS is not in faster convergence, which we found to be only marginal.
The PBRS is, however, substantially more robust than standard shaping, making tuning the reward shaping functions much easier.

\begin{figure}[tb]
  \includegraphics[width=\columnwidth]{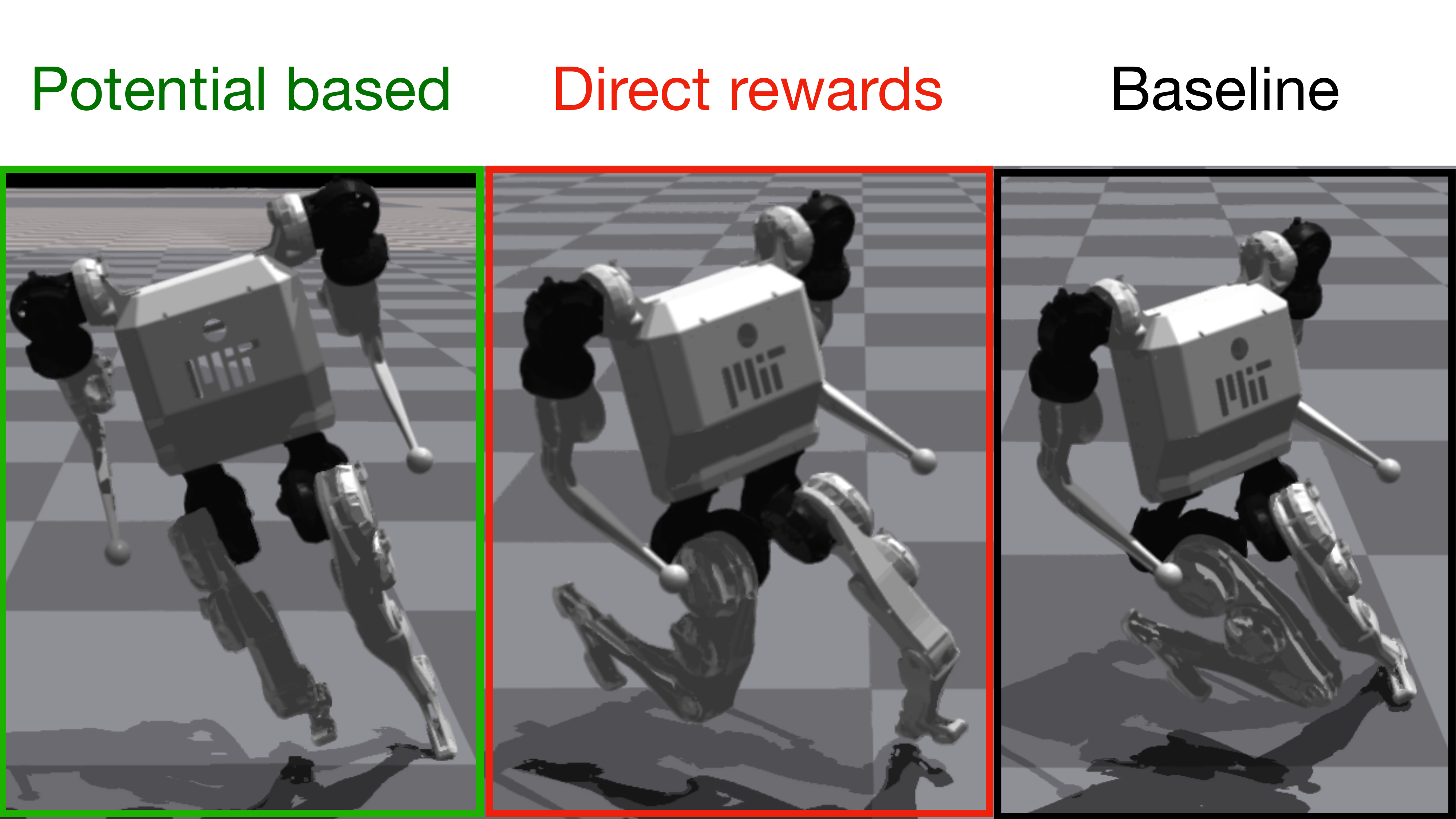}
  \caption{The potential based (left), direct (middle), and baseline (right) locomotion policies. The controller is capable of forward velocities and yaw rates of approximately 3.5 m/s and 1.5 rad/s, respectively.}
  \label{fig:humanoid_policy}
\end{figure}
\subsection{Related Work}
A common approach to provide dense rewards is to directly track references that are highly correlated with the desired behavior.
For example,~\textcite{Peng2020ImitatingAnimals} use motion capture data from dogs to generate reference motion for RL for a quadruped robot, and~\textcite{rai2018BOBiped} use clinical data of humans to obtain reference values such as desired body height and orientation for a bipedal walking robot.
\textcite{green2021learning} pre-compute a library of reference trajectories for a simpler, lower-dimensional system that is amenable to model-based trajectory optimization, and use this library for reward shaping in RL on a bipedal robot.
In a similar fashion,~\textcite{reda2022brachiating} solve for an optimal policy on a simpler model using RL, and use these outputs as shaping rewards for learning brachiation in a simulated 2D animation.
In all these examples, it is critical to carefully choose the references and the weighting of the shaping rewards, as the policy can learn to overfit to the references instead of the intended task.
\par
In an attempt to side-step this problem,~\textcite{Ng_PolicyInvariance} show policy invariance to PBRS (see eq.~\eqref{eq:PBRS}), and recommend using an approximation of the value function.
Using a value function estimate in PBRS form can be seen as performing credit assignment, redistributing the value as instantaneous rewards throughout state-space~\cite[Sec. 4]{Schulman_GAE}.
This is particularly helpful if the baseline reward is sparse, such as a boolean indicator of task-completion.
Indeed,~\textcite{wiewiora2003potential} showed that PBRS is equivalent to initializing Q-values.
Since the optimal policy is greedy with respect to the (true optimal) Q-value function, a well chosen PBRS essentially allows the discount factor to be much more myopic.
~\textcite{westenbroek2022lyapunov} leverage this property for sample-efficient learning directly in hardware, using a discount factor of zero.
Though the reward-shaping is justified with control Lyapunov functions, the main case study on a cartpole uses the value function obtained in simulation and is directly equivalent to PBRS with a value function.
From ablation studies, they also observed that if the value function used is too inaccurate, it is necessary to increase the discount factor.
\par
~\textcite{Harutyunyan_ArbitraryPBRS} and others~\cite{Devlin_DynamicPBRS, Devlin_TheoreticalPBRS, Malysheva_RunWithPBRS} have explored using PBRS in a more general setting, and consistently find that PBRS greatly accelerates convergence on simple problems such as gridworlds or cartpole balancing.
~\textcite{Malysheva_RunWithPBRS} learn locomotion on a higher-dimensional biped constrained to 2D using references similar to those discussed above, but put in PBRS form.
They also report faster convergence, though this is strongly influenced by the quality of the references used.
\par

\subsection{Outline}
In Section II, we review concepts and terms necessary to describe PBRS and its implications for reinforcement learning.
In Section III, we detail the system, observations, and rewards we use for the locomotion task.
In Section IV, we compare the effects of PBRS and DRS for training and on the converged policies.
Lastly, in Section V, we present our conclusions and outline future directions for using PBRS in RL. 

\begin{figure}[tb]
\centering
  \includegraphics[width=0.8\columnwidth]{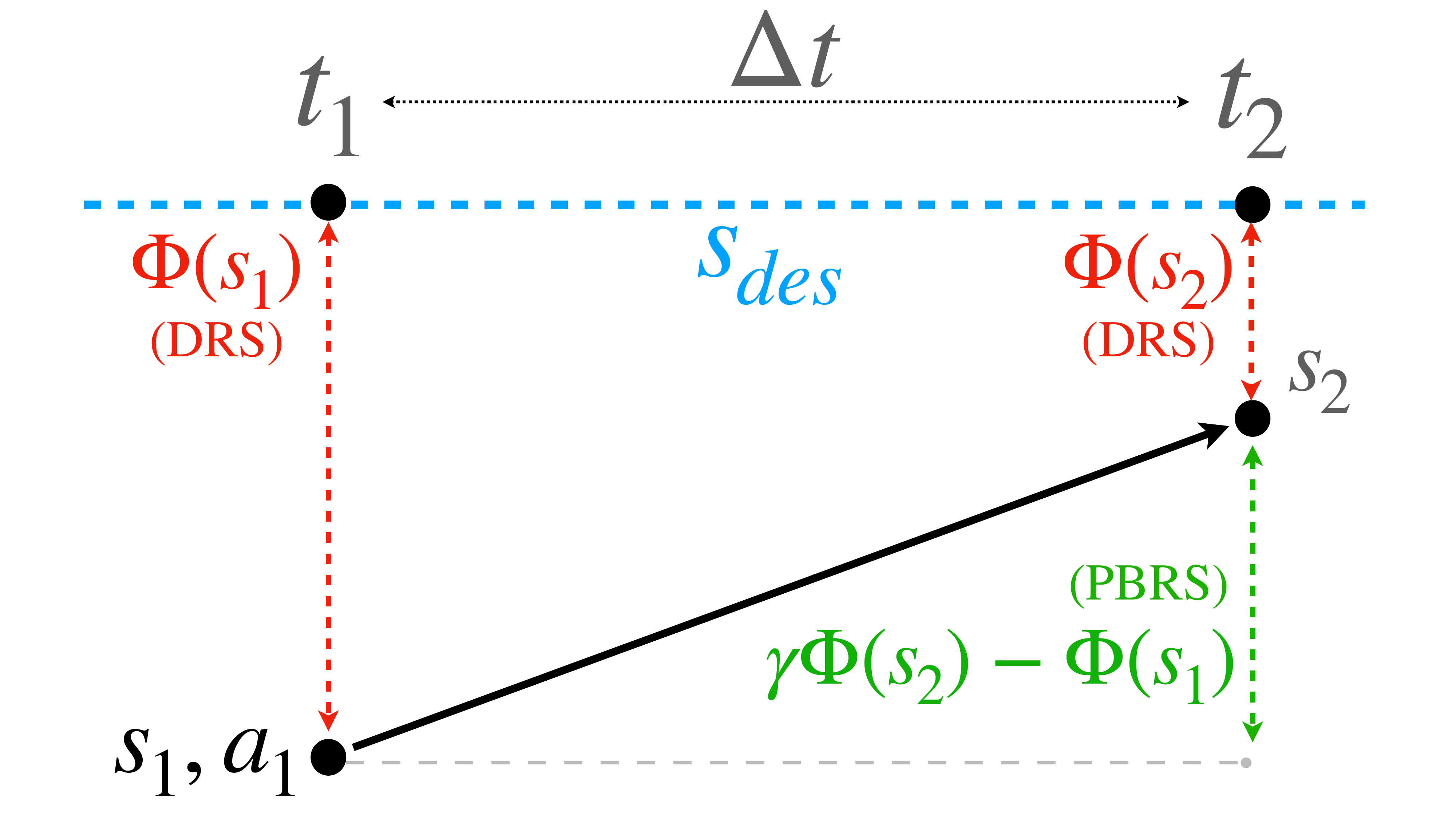}
  \caption{A visualization of a tracking reward in both direct-reward shaping and potential-based reward shaping form. 
  While DRS rewards return the instantaneous evaluation of $\Phi$, PBRS rewards give returns for \textit{improvement} in $\Phi$ at the next state.}
  \label{fig:pbrs_diagram}
\end{figure}

\section{Preliminaries and Problem Statement}
\subsection{Markov Decision Processes and Reinforcement Learning}
A Markov decision process (MDP) is a tuple $\mathcal{M} = \MDP$, with state space $\mathcal{S}$, action space $\mathcal{A}$, transition probabilities $\mathcal{T}(s_{k+1}|s_k, a_k)$ describing the dynamics of the system, reward function $r$ and discount factor $\gamma_t$.
The reward function $r$ takes the general form ${r: \mathcal{S} \times \mathcal{A} \times \mathcal{S} \rightarrow \mathbb{R}}$ with $r(s_k, a_k, s_{k+1})$ describing the reward received for transitioning to state $s_{k+1}$ from $s_k$ with action $a_k$.
\par

For a given MDP, we wish to find a policy $\pi: \mathcal{S} \rightarrow \mathcal{A}$ to maximize the discounted sum of all future rewards.
For a fixed policy, this quantity is represented as the optimal value function, given as 
\begin{equation}
    V^*(s_k) = \mathbb{E}\left[\sum_k^{\infty}\gamma_t^k r(s_k, \pi(s_k), s_{k+1})\right].
\end{equation}

\subsection{Reward Shaping}
Learning the value function is especially challenging when long-term optimality differs strongly from short-term reward signals, for example for sparse, task-based rewards.
In these cases, it is typically necessary to specify a discount factor $\gamma$ close to one, to emphasize the importance of actions on long-term outcomes.
It is more challenging because the agent needs to deal with a more ambiguous credit assignment problem.
A central theme to reward shaping is to provide a reward signal that is more immediately informative about the current action's effect on the final outcome.
This is typically done by formulating \emph{dense} rewards so that informative reward signals are available throughout the trajectory.
\par
We denote these shaping rewards as $\DRS(s_k, a_k, s_{k+1})$, and the corresponding total reward $\hat{r}$ and MDP $\mathcal{M}_{shaped}$ as
\begin{align}
\hat{r}(s_k, a_k, s_{k+1}) &= r(s_k, a_k, s_{k+1}) + \DRS(s_k, a_k, s_{k+1}) \\
\mathcal{M}_{shaped} &= (\mathcal{S}, \mathcal{A}, \mathcal{T}, \hat{r}, \gamma).
\end{align}
When well chosen, shaping rewards can greatly help convergence. 
However, it is important to note that adding shaping terms fundamentally changes the MDP and can have unintended effects on the desired behavior, as discussed by~\textcite{Randlov_BicycleRewardShaping}.
\par
We note that in most RL in robotics studies~\cite{Miki_RLAnymal,green2021learning, Rudin_Parallel, Peng_DeepMimic}, reward terms are typically restricted to functions that can be computed as $r(a)$, and $r(s)$ or $\DRS(s)$. We will slightly abuse notation and refer to $\DRS(s)$ as \emph{direct reward shaping} (DRS).

\subsection{Potential-based Shaping}
The focus of this paper is \emph{potential-based shaping} of the reward function~\cite{Ng_PolicyInvariance}.
Consider a modified MDP, $\mathcal{M}_{potential}=(\mathcal{S}, \mathcal{A}, \mathcal{T}, \tilde{r}, \gamma)$, defined with
\begin{align}\label{eq:PBRS}
\PBRS(s_k, s_{k+1}) &= \gamma \Phi(s_{k+1}) - \Phi(s_k) \\
\tilde{r}(s_k, a_k, s_{k+1}) &= r(s_k, a_k, s_{k+1})+\PBRS(s_k, s_{k+1}),
\end{align}
where $\Phi(\cdot)$ is some scalar, real-valued function and $\PBRS(\cdot)$ is the potential-based reward.
As shown in Fig.~\ref{fig:pbrs_diagram}, potential based rewards are concerned with the \emph{change} of the rewards pushed through the dynamics, as opposed to their instantaneous values.
\par

Note that any DRS term $\DRS(s)$ can trivially be used as a potential function for PBRS, and for the rest of the paper we will focus on comparisons where we use $\Phi(s) = \DRS(s)$.
~\textcite{Ng_PolicyInvariance} presented theoretical results demonstrating that even for arbitrary potential functions, the optimal policy for the original MDP $\mathcal{M}$ is invariant to this class of potential-based shaping rewards.
Furthermore, the advantage function and policy gradients are also unaffected by the addition of $\PBRS(s_k, s_{k+1})$ to the original set of rewards~\cite{Schulman_GAE}.
This means that, in theory, an agent trained on the shaped MDP $\mathcal{M}_{shaped}$ should converge to a policy that is also optimal for the original MDP $\mathcal{M}$.
\par
In practice, however, RL algorithms are affected by a myriad of hyperparameters, such as function approximation choices and exploration heuristics, that prevent the agent from converging to the true optimal policy in reasonable time.
We focus on comparing the performance of shaping with DRS and PBRS in a practical setting for training a humanoid robot to run.

\begin{table}[tb]
\renewcommand*{\arraystretch}{1.5}
    \centering
    \caption{Agent Observations}
    \begin{tabular}{lcc} 
     \textbf{Observation} & \textbf{Dim.} & \textbf{Noise Range} \\ 
     \toprule
      Joint positions $\q$  & 10 & 0.005 \\ 
     \hline
     Joint velocities $\dot{\q}$ & 10 & 0.01 \\ 
     \hline
     Body height $z_b$ & 1 & 0.05 \\ 
     \hline
     Body velocity $\v_b$ & 3 & 0.1  \\ 
     \hline
     Body angular velocity $\mathbf{\omega}_b$ & 3 & 0.05 \\ 
     \hline
     Body frame gravity $\hat{\g}$ & 3 & 0.05 \\ 
     \hline
     Binary foot contact state $\b_c$ & 2 & 0.1 \\ 
     \hline
     Commanded velocities
     $\begin{cases}
        c_x \: (\mathrm{forward})\\
        c_y \: (\mathrm{lateral})\\
        c_\omega \: (\mathrm{yaw})
     \end{cases}$
     & 3 & 0 \\ 
     \hline
     Clock phase
     $\begin{cases}
          \sin(\phi)  \\
          \cos(\phi)\\
          \frac{\sin(\phi)}{2\sqrt{\sin(\phi)^2 + 0.04}} + 0.5
    \end{cases}$
      & 3 & 0 \\
    \end{tabular}
    \label{tab:observations}
\end{table}

\section{Humanoid Locomotion Case Study}
To empirically test the effectiveness of PBRS for continuous, high-dimensional robot control, we benchmark a learning pipeline for running with the MIT Humanoid robot.
We start with a minimal set of \emph{baseline rewards} and then benchmark a set of commonly used DRS reward terms and the same set of shaping rewards reformulated as PBRS reward terms.

\subsection{System Overview}
The MIT Humanoid is an 18 degree-of-freedom robotic platform designed by the Biomimetic Robotics Lab~\cite{chignoli2021humanoid}.
For learning running locomotion, we fix the arm joints at nominal angles and reduce control of the system to only the legs, a total of 10 degrees of freedom.
\par
The policy network consists of a single neural network that outputs joint position targets $\mathbf{a} \in \R^{10}$ to the system, similar to prior work \cite{Siekmann_Stairs, Miki_RLAnymal}.
The torques are calculated as
\begin{equation}
    \boldsymbol{\tau} = K_p(\mathbf{a} - \q) + K_d(\dot{\q}),
\end{equation}
where $K_p = 30$ Nm/rad and $K_d = 5$ Nms/rad are fixed proportional and damping gains respectively, and $\q$ are the joint angles.
\par
The observations $\s \in \R^{38}$ are listed in Table~\ref{tab:observations}, and are affected by uniformly sampled noise.
The phase $\phi$ is a simple clock with constant growth at one Hz, which we found helpful for the policy to settle into a periodic gait, although the final gaits observed are not limited to this frequency.

\begin{table}[tb]
\renewcommand*{\arraystretch}{1.5}
    \centering
    \caption{Training rewards}
    \begin{tabular}{lcc } 
    
     \textbf{Baseline} & \textbf{Weight} & \textbf{Function} \\ 
     \toprule
      Linear velocity  & 10 & \(\exp(-\lvert v_{x,y} - c_{x,y} \rvert ^2/\sigma_{xy}) \) \\ 
     \hline
     Angular velocity & 5 & \(\exp(-(\omega_z - c_\omega)^2/\sigma_\omega) \) \\ 
     \hline
     1st order action rate & -1e-3 & \(\lvert (q_{\pi}^k - q_{\pi}^{k-1})/\Delta t \rvert ^2\) \\ 
     \hline
     2nd order action rate & -1e-4 & \(\lvert (q_{\pi}^k - 2q_{\pi}^{k-1} + q_{\pi}^{k-2})/\Delta t \rvert ^2\)  \\ 
     \hline
     Torques & -1e-4 & \(\lvert \boldsymbol{\tau} \rvert^2\) \\ 
     \hline
     Torque limits & -0.01 & \( \operatorname{max} (\lvert \boldsymbol{\tau} \rvert - \beta_\tau\boldsymbol{\tau}_{max}, 0)\) \\ 
     \hline
     Joint limits & -10 & \( \operatorname{max} (\lvert \boldsymbol{\tau} \rvert - \beta_q\boldsymbol{\tau}_{max}, 0)\) \\ 
     \hline
     Termination & -100 & 
     $\begin{cases}
        1, \; \lvert \mathbf{v}_b \rvert \geq 10 \text{ [m/s]}, \\  
        1, \; \lvert \boldsymbol{\omega}_b \rvert \geq 5 \text{ [rad/s]}, \\
        1, \; \hat{g}_x, \hat{g}_y \geq 0.7, \\
        1, \; \operatorname{self-collision}, \\
        0, \; \operatorname{otherwise}.
     \end{cases}$ \\

    \\
     \textbf{\color{red} Direct Shaping} & \textbf{Weight} & \textbf{Function} \\ 
     \toprule
     Orientation $R_{\mathrm{ori}}$ & 5.0 & $\exp({{-(\hat{g}_x^2 + \hat{g}_y^2)}/{\sigma_\theta}})$ \\
     \hline
     Height $R_{\mathrm{h}}$ & 2.0 & $\exp({{-(z_b - z_{des})^2}/{\sigma_h}})$ \\
     \hline 
     Joint regularization $R_{\mathrm{j}}$ & 1.0 & $\exp({{-(q_a^L - q_a^R)^2}/{\sigma_q}})$ \\
                                             &     & $+ \exp({{-(q_p^L - q_p^R)^2}/{\sigma_q}})$ \\
                                             &     & $+ \exp({{-(q_y^L)^2}/{\sigma_q}})$ \\
                                             &     & $+ \exp({{-(q_y^R)^2}/{\sigma_q}})$ \\
     \\
     \textbf{\color{ForestGreen} Potential Shaping} & \textbf{Weight} & \textbf{Function} \\ 
     \toprule
     Orientation & 1.0 & $\gamma R_{\mathrm{ori}}(\mathbf{s}_{k+1}) - R_{\mathrm{ori}}(\mathbf{s}_k)$ \\
     \hline
     Height & 1.0 & $\gamma R_{\mathrm{h}}(\mathbf{s}_{k+1}) - R_{\mathrm{h}}(\mathbf{s}_k)$ \\
     \hline 
     Joint regularization & 1.0 & $\gamma R_{\mathrm{j}}(\mathbf{s}_{k+1}) - R_{\mathrm{j}}(\mathbf{s}_k)$\\
     \end{tabular}
    \label{tab:humanoid_baseRewards}
\end{table}


\subsection{Baseline Rewards}
For general locomotion, we define a set of baseline rewards as in Table~\ref{tab:humanoid_baseRewards}.
The first two, linear velocity and angular velocity tracking, are the only task-related rewards; all other terms are generic regularization terms to encourage smoothness, efficiency, and discourage joint limit violations.
\par
For the baseline rewards, ${\Delta t}$ is the controller timestep, ${\sigma=0.5}$ is a scaling parameter, ${\beta_\tau=0.8}$ and ${\beta_q=0.9}$ act as soft-stop limits to discourage reaching the joint and actuator constraints, and ${\boldsymbol{\tau}_{max}}$ and $q_{max}$ are torque and joint limits of the system respectively.


\begin{table}[tb]
    \renewcommand*{\arraystretch}{1.2}
    \centering
    \caption{Training Environment Hyperparameters}
    \begin{tabular}{lc } 
     \textbf{Hyperparameter} & \textbf{Value} \\ 
     \toprule
     Value loss coefficient & 1.0 \\ 
     \hline
     Clipping $\epsilon$ & 0.2 \\ 
     \hline
     Entropy coefficient  & 0.1 \\ 
     \hline
     Learning rate & 1e-5 (adaptive)  \\ 
     \hline
     Discount factor & 0.99  \\ 
     \hline
     $\lambda$ & 0.95 \\ 
     \hline
     Steps/env & 24 \\ 
     \hline
     Policy network size & [256, 256, 256]\\ 
     \hline
     Critic network size & [256, 256, 256]\\ 
     \hline
     Activation & ELU\\ 
     \hline
    \end{tabular}
    \label{tab:humanoid_hyperparameters}
\end{table}

\subsection{Shaping Rewards}
We choose three shaping rewards commonly used as costs in the humanoid locomotion literature~\cite{Dai_WBPlanning, Garcia_HumanoidMPC, rai2018BOBiped}:
we regularize the \emph{orientation} ($R_{\mathrm{ori}}$), \emph{height} ($R_{\mathrm{h}}$), and \emph{joints} ($R_{\mathrm{j}}$), with their respective reward terms defined in Table \ref{tab:humanoid_baseRewards}.
A nominal desired height $z_{des} = 0.6$ m is a hand-chosen height target, $\hat{g}_x$, $\hat{g}_y$ are the components of the gravity vector in the body frame, and $q_i^j$ is the $i^{th}$ joint type on leg $j$.
The subscript denotes the specific joint, with $q_a$, $q_p$, and $q_y$ representing the abduction/adduction, pitch, and yaw joints specifically, and the superscript refers to the leg (left/right) the joint is part of.
The reward $\DRS_{joint}$ serves to regularize the yaw joints about zero and encourage symmetry between the ab/ad and pitch joints of the legs.
We use squared-exponential functions to define our reward functions, as is common in RL literature~\cite{Miki_RLAnymal, Peng_DeepMimic, Peng2020ImitatingAnimals, Siekmann_Stairs}.
\par
We can put these "direct" shaping rewards in their potential based forms trivially as
\begin{align}
    \PBRS_{\mathrm{s}}(\mathbf{s}_k, \mathbf{s}_{k+1}) &= \gamma \DRS_{\mathrm{s}}(\mathbf{s}_{k+1}) - \DRS_{\mathrm{s}}(\mathbf{s}_k),
\end{align}
for some arbitrary shaping reward $R_{\mathrm{s}}$ and potential discounting $\gamma$. 

\subsection{Implementation Details}
The locomotion policy is trained in the NVIDIA IsaacGym framework open-sourced by~\textcite{Rudin_Parallel} with the PPO-Clip algorithm~\cite{Schulman_PPO} and the hyperparameters shown in Table~\ref{tab:humanoid_hyperparameters}.
The agents are trained on a computer equipped with an Intel i9-10850K processor and NVIDIA RTX 3060 GPU.
The simulation is run at 1000 Hz, with a control frequency of 100 Hz. Each training run includes 4096 agents and is run for 1000 policy iterations, and we see convergence within around 30 minutes of wall-clock time.
\par

\begin{figure}[tb]
\centering
  \includegraphics[width=0.8\columnwidth]{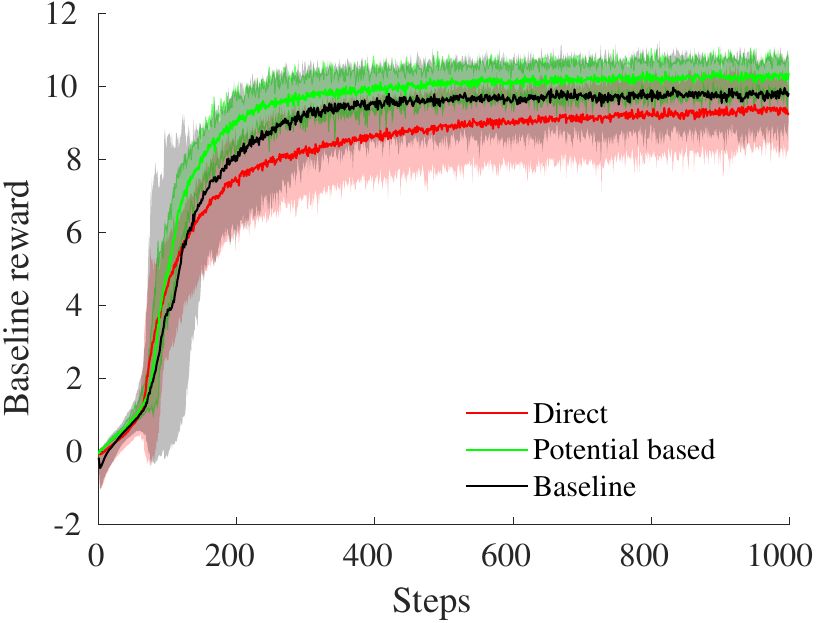}
  \caption{Values for the total baseline rewards during training for the PBRS, DRS, and baseline policies.}
  \label{fig:rwd_benchmark}
\end{figure}

\begin{figure*}[tb]
\centering
\begin{subfigure}{.45\textwidth}
  \centering
  \includegraphics[width=1.\linewidth]{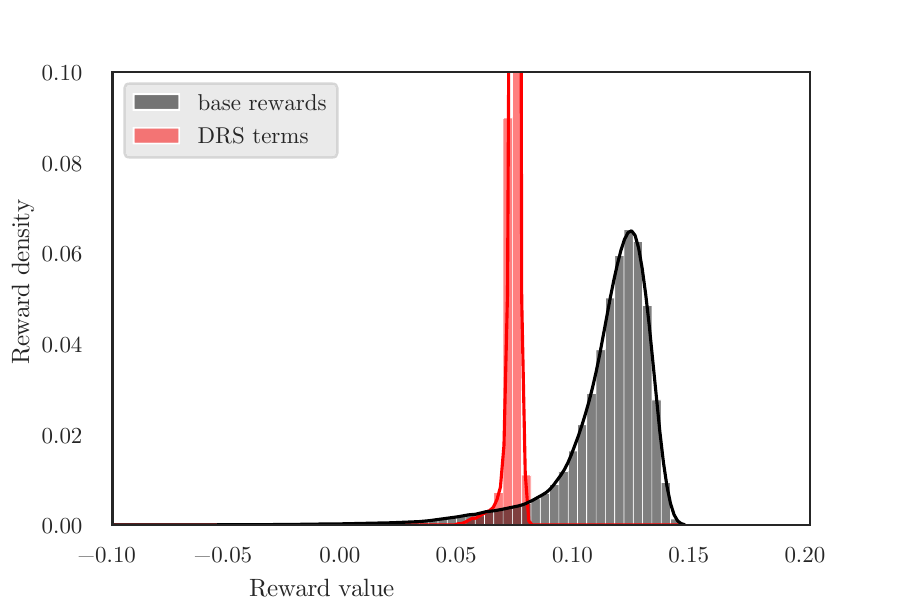}
  \caption{Distribution at iteration 500}
  \label{fig:sfig1}
\end{subfigure}%
\begin{subfigure}{.45\textwidth}
  \centering
  \includegraphics[width=1.\linewidth]{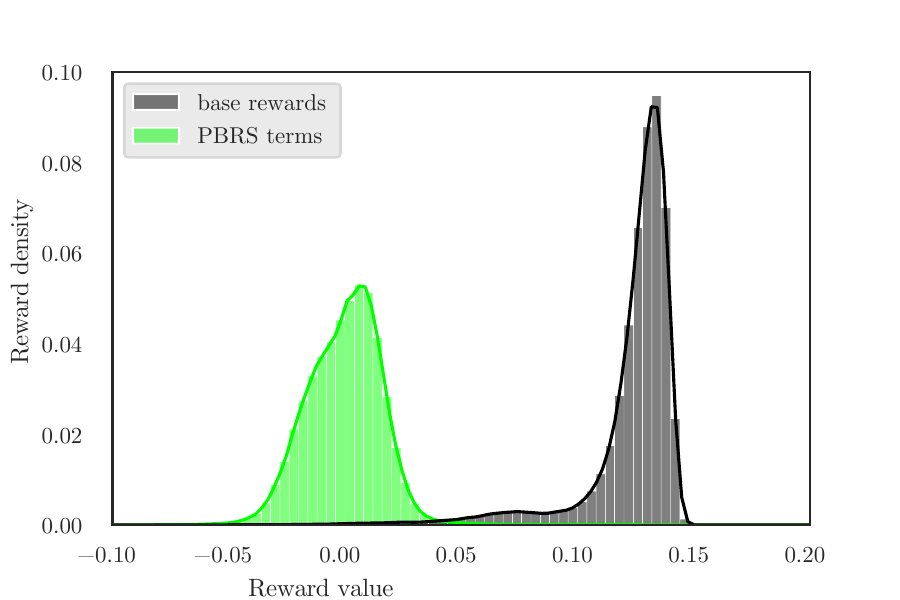}
  \caption{Distribution at iteration 500}
  \label{fig:sfig2}
\end{subfigure}
\caption{Distribution of rewards during an iteration at iteration 500 (well converged).
Although the distributions of rewards begins rather spread out, agents tend to quickly fit to dense DRS terms (\ref{fig:sfig1}), whereas the distribution of PBRS terms remains centered around zero and with a relatively wide distribution throughout training (\ref{fig:sfig2}).
See accompanying video for the evolution of this distribution during training.}
\label{fig:distribution}
\end{figure*}

\section{Results}
We present here a benchmark with three cases: baseline rewards only, baseline rewards with DRS terms, and baseline rewards with PBRS terms, with accompanying video results and code\footnote{Video: https://youtu.be/Qvacov9kujQ}\footnote{Code: https://github.com/se-hwan/pbrs-humanoid}.
We first tune the baseline reward weights until reasonable running performance is achieved (see Table~\ref{tab:humanoid_baseRewards}), then keep those weights fixed for all experiments.
The weights of the DRS and PBRS rewards are then tuned until a reasonable locomotion policy is found, with the weights set to the tuned values in Table \ref{tab:humanoid_baseRewards}.
Cases are compared with accumulated baseline rewards and not the total rewards, such that the comparisons are not affected by the scaling of shaping rewards.
We also visually inspect policies to evaluate the resulting behavior for qualitative differences. 

Trajectories are collected over 24 timesteps, equating to roughly 0.2 s of simulation time, and the agents are subjected to randomized impulses, friction, and velocity commands during training.

\subsection{Discounting of Potential-Based Rewards}
We find that in practice, using discounting for PBRS in~\eqref{eq:PBRS} can lead to learning instability~\cite{Grzes2009AnalysisPBRS}.
To overcome this issue, we set $\gamma=1$ in the calculation of~\eqref{eq:PBRS} (though not in calculating the advantage for PPO).
While policy invariance is technically sacrificed by doing so, we find that training converges far more stably and quickly with this modification.
We do not discuss it further as our findings to this regard closely match those of~\textcite{Grzes2009AnalysisPBRS}, who studied in detail how the discount factor in potential-based rewards can affect both the magnitude and sign of the returned value.
We confirm their finding, as other studies on PBRS do not report any modifications of the discounting~\cite{Devlin_DynamicPBRS, Devlin_TheoreticalPBRS, Malysheva_RunWithPBRS}, yet we found this to make a significant difference in the effectiveness of PBRS terms.

\begin{figure}[bt]
 \centering
  \includegraphics[width=0.8\columnwidth]{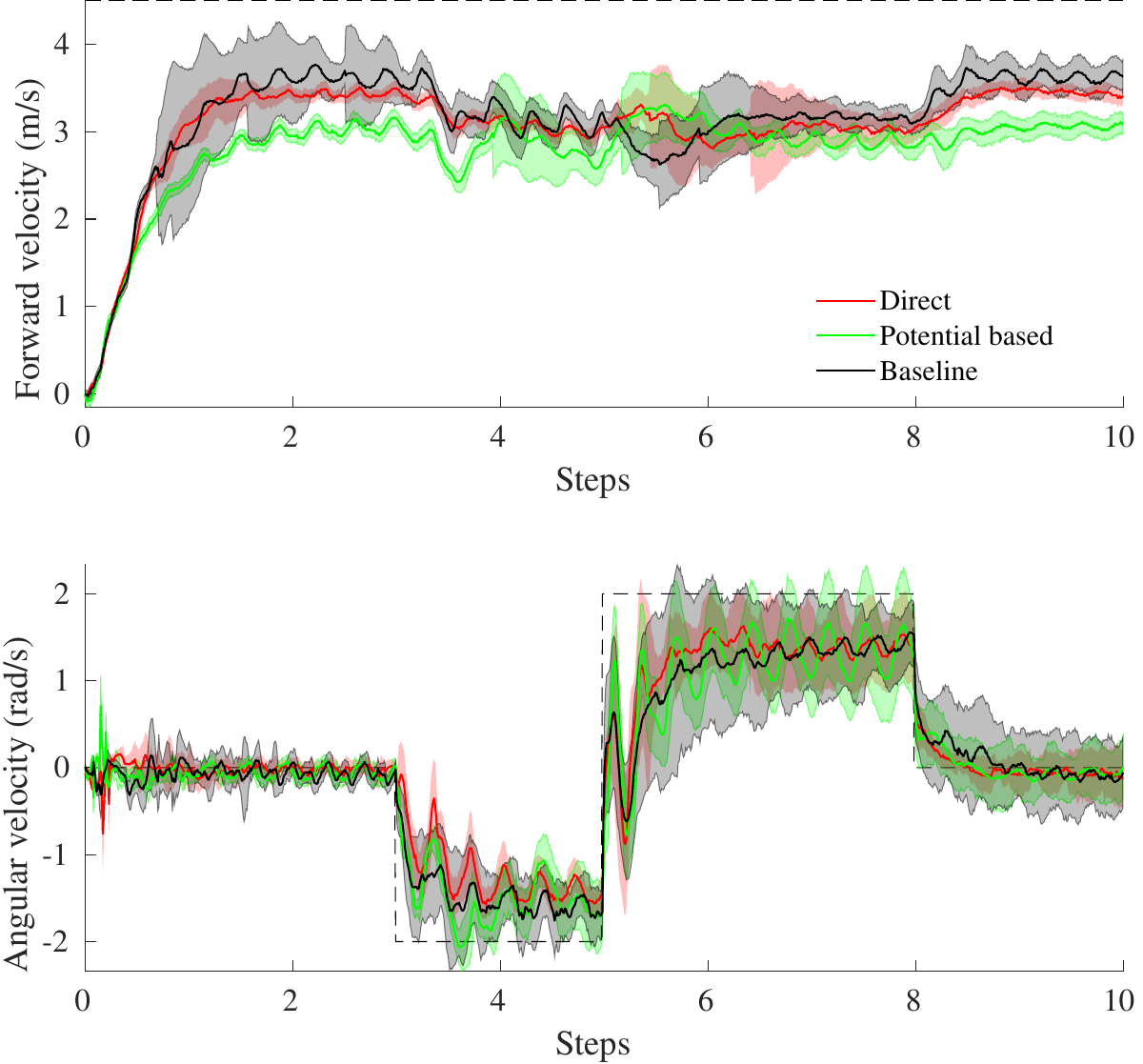}
  \caption{Velocity tracking plots for the PBRS, DRS, and baseline policies over 50 runs.
  The dotted line indicates the commanded forward velocity (top) and yaw angular rate (bottom), respectively.}
  \label{fig:humanoid_performance}
\end{figure}

\subsection{Training Benchmark}
We run the three cases (baseline, DRS, and PBRS) ten times each with the tuned, nominal weights for all rewards.
All three cases converge to behaviors with similar performance within 1000 iterations, as seen in the learning curves shown in Fig.~\ref{fig:rwd_benchmark}.
The agent with PBRS terms converges to a policy with slightly higher return, slightly more quickly, though this improvement is relativey marginal.
The variance between the ten learning runs, however, is significantly lower, finishing at 0.946, compared to 1.905 when using DRS terms, and 2.025 when using baseline rewards only.
\par

\begin{figure*}[tb]
\centering
  \includegraphics[width=0.9\textwidth]{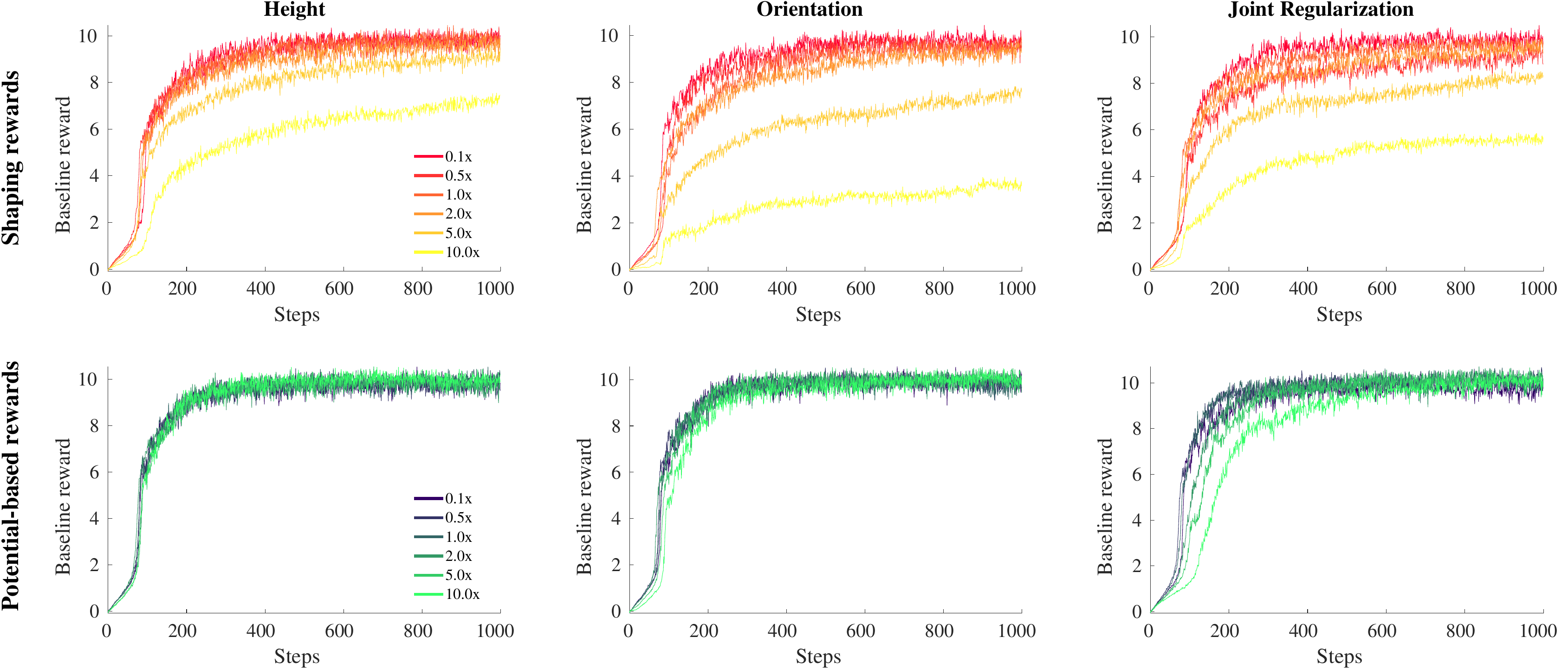}
  \caption{To compare the robustness of using potential-based shaping against direct reward shaping, we train policies with the baseline rewards in combination with either the height, orientation, or joint regularization reward.
  From a tuned nominal value, we then sweep across the weights of that reward from 0.1$\times$ to 10$\times$ the nominal value.
  The potential-based rewards were far less sensitive to large changes in scaling compared to the shaping rewards.}
  \label{fig:humanoid_sweep}
\end{figure*}

Agents trained with DRS terms converge to policies with slightly lower returns than the baseline; however, when inspecting the policies, we find that baseline policies tend to have abnormal gaits, with legs turned inward.
The most likely explanation for this is the greater mediolateral stability it provides, which assists with avoiding termination early on in training.
This behavior persists and appears detrimental when turning at high velocities, as shown in Figs.~\ref{fig:humanoid_policy} and~\ref{fig:humanoid_performance}.
Both DRS and PBRS terms rectify this issue by regularizing the yaw joints around zero.

In the case of DRS, however, the agent appears to strongly prioritize maximizing the shaping reward terms, which likely conflicts with maximizing the baseline reward.
This can be seen when inspecting the reward distribution during training, shown in Fig.~\ref{fig:distribution}: the PBRS terms remain centered around 0 and maintain a relatively large spread, even halfway through training, whereas the DRS terms are very quickly maximized, which suggests the agent prioritizes maximizing the shaping rewards.
This empirical observation supports the theory behind the policy invariance of PBRS.
Because PBRS terms become zero-mean centered as training progresses, their influence on the optimizer decreases, allowing greater returns on the baseline rewards.

\par
Another interesting result is the emergence of natural heel-toe transitions for the trained policies.
While none of the rewards explicitly specify this behavior, the touchdown and push-off phases of stance are quite clear.
It is possible that the relatively low $K_p$ gains on the joints force the robot into this pattern, but isolating the rewards and environment configurations that can reproduce this behavior is beyond the scope of this work.
We present these details as examples of how desirable policies can be both difficult to express and sometimes counterintuitive. 

To evaluate the policy itself, we compare the linear and angular velocity tracking performance of the three policies at the limits of the commands, as shown in Fig.~\ref{fig:humanoid_performance}.
We find no significant difference in command tracking on average between the DRS and PBRS policies, but observed that the baseline policy often terminates during the turns, which accounts for the large standard deviation in angular velocity between $t = 3$ and $t=10$ s in Fig.~\ref{fig:humanoid_performance}.
While the DRS policy tracks the desired velocity most closely, this metric alone does not account for all the other terms defined as part of the "baseline reward", which overall, is higher for the potential-based policy.
\par
We also compare the average base heights of the DRS and PBRS through the trajectory.
As expected, the average height of the DRS policy is 0.596 m, close to the specified desired height of 0.60 m.
However, with the same reward used in a potential-based form, the average height of the PBRS is 0.639 m, corresponding to a change of almost 5\% of the total height of the robot. 
While both PBRS and DRS learn comparable policies, the policy appears to be more strongly biased by the DRS terms than PBRS terms.
By formulating these rewards in their potential based forms, we retain the advantages of being able to guide the policy towards desirable states while relaxing how much it is affected by the rewards.

\subsection{Sensitivity Analysis}
We perform a sensitivity analysis of both DRS terms and PBRS terms by sweeping from 0.1 to 10 times the nominal weights.
As shown in Fig.~\ref{fig:humanoid_sweep}, learning with PBRS terms is substantially more consistent across the weights and individual shaping rewards chosen.
When the weights of the DRS terms are increased, the policy overfits to the specified reward and sacrifices the performance of the baseline rewards to do so.
In particular, rewarding a fixed, upright orientation is particularly detrimental to the baseline rewards.
This is unsurprising, given the significant banking and oscillations of the torso that naturally occur during running motions. 

By placing shaping rewards in potential based form, the range of weights that can produce desirable behavior is much larger.
This significantly eases the burden of iterating on sets of reward weights for an acceptable policy.
\section{Conclusion and Outlook}
We find that PBRS terms are beneficial for learning on high-dimensional, continuous systems such as legged robots; unlike previous studies~\cite{Devlin_DynamicPBRS, Malysheva_RunWithPBRS, westenbroek2022lyapunov}, which have mostly focused on gridworld or low-dimensional systems, we find that the main benefit is not in accelerated convergence (which in our case is only marginal) but rather on ease of tuning.
We note that RL implementations in robotics often only use rewards of the form $r(s_k)$; this type of reward can be trivially converted into PBRS form, and from our findings, we advocate using the PBRS form of rewards when possible.

While we found that PBRS terms are relatively robust to weighting, we also note that these terms are implicitly scaled through the dynamics by the control timescale $\Delta t$.
In future work, we plan to more closely investigate this relationship, especially in the context of hierarchical RL.
Since control timescales are a natural approach to choosing hierarchical levels (typically, higher levels in a hierarchy will reason on a longer horizon, with a larger $\Delta t$), it may be possible to automatically assign rewards for different tasks to different hierarchy levels based on their relative scaling.
\par
Another promising avenue in the context of hierarchical RL is to use value functions as PBRS, as originally proposed by~\textcite{Ng_PolicyInvariance}.
Although finding a good approximation for a value function is often daunting, in a hierarchy of world models, it may be possible to solve a cascade of problems using a hierarchy of simplified world-models, and use the obtained value function to shape the reward of each successive stage.

\printbibliography
\end{document}